\documentclass{article}

\usepackage[dblblindworkshop, final]{neurips_2026}

\usepackage[utf8]{inputenc} 
\usepackage[T1]{fontenc}    
\usepackage{hyperref}       
\usepackage{url}            
\usepackage{booktabs}       
\usepackage{amsfonts}       
\usepackage{nicefrac}       
\usepackage{microtype}      
\usepackage{xcolor}         

\usepackage{graphicx}
\usepackage{bbm}
\usepackage[ruled,vlined]{algorithm2e}
\usepackage{amsmath}
\usepackage{natbib}

\title{Progressive Subset Selection and Cascaded Consistency for Unsupervised Vision–Language Model Adaptation}
\workshoptitle{NeurIPS 2026 Workshop on Continual Learning in the Era of Foundation Models and Embodied Agents}

\author{%
  Hao Yan,~ Guanghui Wang \\
  Department of Computer Science \\
  Toronto Metropolitan University \\
  Toronto, ON, Canada \\
  \texttt{haoyan6@cmail.carleton.ca, wangcs@torontomu.ca}
}

\begin{document}

\maketitle

\begin{abstract}
  Pretrained Vision-Language Models (PVLMs) achieve strong zero-shot recognition but often suffer performance degradation under target-domain shift. Unsupervised adaptation provides a practical way to improve target-domain performance without labeled data, but it remains challenging because pseudo-labels generated by the model can be unreliable and no labeled validation set is available for model selection or early stopping. We propose a simple framework for unsupervised PVLM adaptation that combines progressive margin-based subset selection with cascaded consistency learning. The subset selection strategy constructs a class-balanced pseudo-labeled set using prediction margins and gradually expands it during training, enabling adaptation to start from reliable supervision anchors before incorporating broader target-domain data. Cascaded consistency learning further exploits unlabeled samples by using temporally smoothed teacher predictions on weakly augmented views to supervise student predictions on strongly augmented views, with confidence filtering to reduce the influence of uncertain pseudo-labels. To preserve the pretrained embedding structure, the framework updates only normalization parameters in the visual encoder and class prototypes initialized from text prompts. Experiments across multiple recognition benchmarks with both CLIP and SigLIP backbones show consistent improvements over zero-shot baselines and competitive performance against recent unsupervised PVLM adaptation methods. These results suggest that progressive pseudo-label selection and teacher-guided consistency provide an effective and lightweight strategy for adapting PVLM classifiers without labeled target data.
\end{abstract}

\section{Introduction}
\label{sec:intro}

Pretrained vision-language models (PVLMs), such as CLIP \citep{radford2021learning}, achieve strong zero-shot recognition by aligning images and text in a shared embedding space. With natural language prompts as class descriptions, these models can recognize diverse categories without task-specific training \citep{radford2021learning,jia2021scaling,li2022blip}. However, their performance often drops when target data differ from the pretraining distribution \citep{radford2021learning,shu2022test}. This motivates adaptation methods that use either limited labeled data through prompt learning \citep{zhou2022learning,zhou2022conditional} or unlabeled target data through unsupervised adaptation \citep{mirza2023lafter,ali2025dpa}.

Unsupervised adaptation is attractive when labels are unavailable, but it is difficult because training relies on pseudo-labels generated by the model itself \citep{lee2013pseudo,arazo2020pseudo}. Under domain shift, these pseudo-labels can be noisy, especially early in training. Optimizing directly on unreliable pseudo-labels may reinforce incorrect predictions and cause unstable adaptation \citep{arazo2020pseudo}. This issue is important for PVLMs because zero-shot classifiers are built from image--text similarity, and high-confidence predictions may still be wrong when the target visual distribution shifts. Since labeled validation data is unavailable, unsupervised adaptation also lacks reliable model selection or early stopping. A practical method should therefore improve target-domain performance while limiting the effect of noisy pseudo-labels. This motivates a conservative adaptation strategy that starts from more reliable predictions and gradually incorporates broader target-domain supervision.

To address this challenge, we propose a simple framework for unsupervised PVLM adaptation that combines progressive margin-based subset selection with cascaded consistency learning. First, we select a class-balanced subset of high-confidence pseudo-labeled samples using prediction margins and progressively expand this subset during training. This allows the model to begin adaptation from reliable supervision anchors before incorporating more target samples. Second, we use cascaded consistency learning, where an exponential-moving-average teacher provides predictions on weakly augmented views to supervise student predictions on strongly augmented views. A confidence threshold filters uncertain teacher predictions, reducing the impact of noisy pseudo-labels. Together, these two components separate reliable direct supervision from broader consistency regularization over the full unlabeled target set.

The proposed framework combines ideas that are widely used in semi-supervised learning and domain adaptation, including pseudo-label filtering, progressive self-training, EMA teachers, and weak-to-strong consistency. Our contribution is to adapt and couple these mechanisms for unsupervised PVLM adaptation, where pseudo-labels are produced by a zero-shot image--text classifier under domain shift and no labeled validation data is available. To avoid disrupting the pretrained embedding space, we update only normalization parameters in the visual encoder and class prototypes initialized from text prompts.

Experiments on multiple benchmark datasets show that the proposed method consistently improves over zero-shot baselines across both CLIP and SigLIP backbones. Compared with recent unsupervised PVLM adaptation methods, the method achieves competitive performance and performs particularly well on datasets with larger zero-shot domain gaps, such as FGVCAircraft and DTD. While it does not outperform all prior methods on every dataset, the results suggest that progressive pseudo-label subset selection and cascaded consistency provide an effective strategy for adapting PVLM classifiers without labeled target data.

Our contributions are summarized as follows:
\begin{itemize}
\item We propose a simple framework for unsupervised vision-language model adaptation that combines progressive margin-based subset selection with cascaded consistency learning.
\item We introduce a progressive pseudo-labeled subset selection strategy that constructs class-balanced supervision anchors using margin-based confidence estimation.
\item We develop a cascaded consistency learning mechanism that combines temporally stabilized teacher predictions, weak-to-strong consistency, and confidence filtering to reduce noisy supervision.
\item We evaluate the framework across CLIP and SigLIP architectures, showing consistent improvements over zero-shot baselines and competitive performance against recent unsupervised PVLM adaptation methods.
\end{itemize}



\subsection{Related Work}\label{sec:related}

\subsubsection{Pretrained Vision--Language Models}

Pretrained vision--language models (PVLMs) learn aligned visual and textual representations from large-scale image--text pairs \citep{asfour2026distilling}. CLIP \citep{radford2021learning} shows that contrastive image--text pretraining enables strong zero-shot recognition by matching image features with textual class descriptions. Subsequent models improve this paradigm through larger datasets, stronger architectures, or alternative training objectives, including ALIGN \citep{jia2021scaling}, OpenCLIP \citep{cherti2023reproducible}, EVA-CLIP \citep{sun2023eva}, and SigLIP \citep{zhai2023sigmoid}. SigLIP replaces the softmax contrastive objective with a sigmoid-based loss, making it useful for evaluating whether adaptation methods generalize beyond CLIP-style models. Despite their strong zero-shot ability, PVLMs can degrade under distribution shift, motivating adaptation methods for target domains.

\subsubsection{Adaptation of Vision--Language Models}

A common way to adapt PVLMs is prompt learning, which optimizes task-specific prompts while keeping most of the pretrained model fixed. CoOp \citep{zhou2022learning} and CoCoOp \citep{zhou2022conditional} learn continuous prompts from labeled examples, while later methods explore improved prompt representations and knowledge-guided prompting \citep{yao2023visual,li2025pf3det,liu2025progressive}. These methods are effective but require labeled data.

Unsupervised PVLM adaptation aims to improve target-domain performance without labeled target samples. CuPL \citep{pratt2023does} uses language models to generate richer class descriptions for zero-shot recognition. UPL \citep{huang2022unsupervised} and POUF \citep{tanwisuth2023pouf} optimize prompts or fine-tune models using unlabeled data. Pseudo-label-based methods such as LaFTer \citep{mirza2023lafter} and DPA \citep{ali2025dpa} adapt PVLMs using model predictions as supervision, with DPA improving pseudo-label reliability through distribution-aware prototype alignment. More recent approaches also study unsupervised knowledge distillation and test-time adaptation, including PromptKD \citep{li2024promptkd}, KDPL \citep{mistretta2024improving}, and TPT \citep{shu2022test}.

\subsubsection{Pseudo-Label Learning and Consistency Regularization}

Pseudo-labeling uses model predictions as supervision for unlabeled data \citep{lee2013pseudo}, but noisy pseudo-labels can cause confirmation bias and unstable training \citep{arazo2020pseudo}. To reduce this risk, prior work often filters pseudo-labels using confidence, uncertainty, entropy, or margin-based criteria. Consistency regularization is another common strategy, encouraging predictions to remain stable under input perturbations. Mean Teacher \citep{tarvainen2017mean} uses an exponential-moving-average teacher to generate stable targets, while FixMatch \citep{sohn2020fixmatch} combines confidence filtering with weak-to-strong augmentation consistency.

Our method builds on these established ideas and adapts them to unsupervised PVLM adaptation. In this setting, pseudo-labels come from a zero-shot image--text classifier under domain shift, and no labeled validation set is available for checkpoint selection. We therefore select class-balanced pseudo-labeled anchors using prediction margins, progressively expand the selected subset, and combine these anchors with EMA teacher-guided weak-to-strong consistency. The contribution is not margin filtering or teacher-student consistency alone, but their combination as a conservative adaptation procedure for PVLM classifiers with unlabeled target data.


\section{Proposed Method}
\label{sec:method}

\begin{figure}[t]
    \centering
    \includegraphics[width=\linewidth]{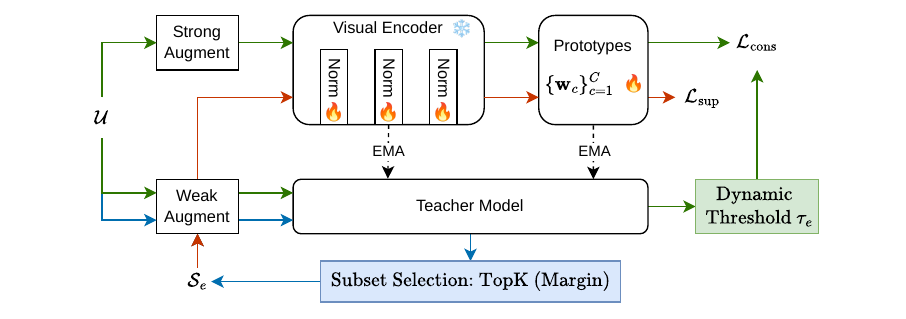}
    \caption{
    Overview of the proposed unsupervised vision-language model adaptation framework. 
    The teacher model selects a class-balanced pseudo-labeled subset based on prediction margins. 
    The student is trained with supervised learning on the selected subset and cascaded consistency learning on unlabeled data via weak and strong augmentations. 
    The teacher parameters are updated as an exponential moving average of the student parameters, producing stabilized pseudo-labels that progressively improve subset selection and consistency learning.
    }
    \label{fig:framework}
\end{figure}

We present a simple framework for unsupervised adaptation of pretrained vision-language models. The framework is model-agnostic and applies to a wide range of pretrained vision-language models (PVLMs) that learn aligned image-text embedding spaces. As illustrated in Figure~\ref{fig:framework}, the method has two complementary components: \emph{progressive margin-based subset selection} and \emph{cascaded consistency learning}. The first component constructs a reliable pseudo-labeled subset by selecting class-balanced samples with large prediction margins and progressively expanding the subset during training to increase supervision coverage. The second component exploits the full unlabeled dataset through cascaded consistency that combines temporally stabilized teacher predictions with cross-view consistency under stochastic augmentations. Together, they form a self-reinforcing loop: reliable pseudo-labeled anchors stabilize early adaptation, while consistency regularization improves representation quality and further enhances pseudo-label reliability. 

Rather than introducing margin filtering or teacher-student consistency as isolated new techniques, our goal is to adapt and combine these established mechanisms for the specific setting of unsupervised PVLM adaptation, where pseudo-labels are produced by a zero-shot image-text classifier under domain shift. To preserve the pretrained embedding geometry and maintain stable optimization under noisy supervision, we adopt a lightweight adaptation strategy that updates only normalization parameters of the visual encoder and class prototypes derived from text prompts.


\subsection{Preliminaries and Adaptation Framework}
\label{sec:prelim}

We consider unsupervised adaptation of a pretrained vision-language model (PVLM) to a target domain without labeled supervision. Modern PVLMs, such as CLIP and SigLIP, consist of two encoders trained on large-scale image-text pairs: a visual encoder and a textual encoder that project images and text prompts into a shared $d$-dimensional embedding space. We denote an input image by $\mathbf{x}$ and a text prompt by $\mathbf{t}$. The visual encoder $\mathcal{V}(\cdot)$ maps $\mathbf{x}$ to $\mathbf{v}=\mathcal{V}(\mathbf{x})\in\mathbb{R}^{d}$, and the textual encoder $\mathcal{T}(\cdot)$ maps $\mathbf{t}$ to $\mathbf{e}=\mathcal{T}(\mathbf{t})\in\mathbb{R}^{d}$. 
During pretraining, these models align image and text embeddings in a shared representation space using similarity-based objectives:
\begin{equation}
\mathrm{sim}(\mathbf{x}, \mathbf{t})
=
\alpha\,
\frac{\mathbf{v}}{\|\mathbf{v}\|_2}
\cdot
\frac{\mathbf{e}}{\|\mathbf{e}\|_2}.
\end{equation}
Here $\alpha$ is a learned temperature/scale that controls the magnitude of similarity logits. This induces a metric structure where semantically aligned image-text pairs exhibit high cosine similarity.

For a downstream classification task with $C$ categories, each class $c\in\{1,\dots,C\}$ is described by multiple textual prompts to reduce prompt sensitivity and capture semantic variability. Let $\{\mathbf{t}_{c,m}\}_{m=1}^{M}$ denote the prompts for class $c$. We compute the normalized embedding of each prompt and average them to form a class prototype:
\begin{equation}
\mathbf{w}_c
=
\frac{1}{M}
\sum_{m=1}^{M}
\frac{\mathcal{T}(\mathbf{t}_{c,m})}{\|\mathcal{T}(\mathbf{t}_{c,m})\|_2}.
\end{equation}
Averaging normalized prompt embeddings mitigates variance from prompt wording and yields a robust semantic prototype. After initialization, $\{\mathbf{w}_c\}_{c=1}^{C}$ are treated as learnable parameters and updated by gradient descent; the text encoder is not used thereafter. When multiple prompt sets are available, the same prompt construction is used consistently for the zero-shot model and the adapted model to isolate the effect of adaptation.

We build a cosine classifier on top of the visual encoder using cosine similarity between image features and class prototypes. Let $\mathcal{V}_{\theta_v}(\cdot)$ denote the visual encoder with parameters $\theta_v$. For an input image $\mathbf{x}$, we compute a normalized visual feature
\[
\bar{\mathbf{v}}
=
\frac{\mathcal{V}_{\theta_v}(\mathbf{x})}{\|\mathcal{V}_{\theta_v}(\mathbf{x})\|_2}.
\]
The overall classifier parameters are
$\theta = \{\theta_v, \{\mathbf{w}_c\}_{c=1}^{C}\}$,
including the visual encoder parameters and class prototypes.
The cosine classifier produces logits $\mathbf{z}=f_\theta(\mathbf{x})\in\mathbb{R}^{C}$, where the $c$-th logit is
\begin{equation}
f_\theta(\mathbf{x})_c
=
\alpha\,
\bar{\mathbf{v}}^\top
\bar{\mathbf{w}}_c,
\qquad
\bar{\mathbf{w}}_c
=
\frac{\mathbf{w}_c}{\|\mathbf{w}_c\|_2}.
\end{equation}
We keep $\alpha$ fixed during adaptation. This cosine classifier preserves the geometric structure of the pretrained embedding space while enabling task-specific discrimination through prototype refinement.

Let $\mathcal{U}=\{\mathbf{x}_i\}_{i=1}^{N}$ denote an unlabeled target dataset from a shifted distribution. The goal of unsupervised adaptation is to improve performance of $f_\theta$ on this target domain without access to labeled target samples. In contrast to supervised fine-tuning, unsupervised adaptation lacks labeled validation data for model selection or early stopping, making training sensitive to noisy pseudo-labels and prone to confirmation bias and late-stage degradation.

To preserve the pretrained geometry and ensure efficient adaptation, we adopt a lightweight update strategy: we update only the affine parameters of normalization layers within $\mathcal{V}_{\theta_v}$ and the class prototypes $\{\mathbf{w}_c\}_{c=1}^{C}$, while keeping the remaining encoder weights fixed. Updating normalization layers adjusts feature statistics to better match the target distribution, and refining prototypes adapts decision boundaries with minimal disruption. This design is conservative: it limits the number of trainable parameters and reduces the risk that noisy pseudo-labels will substantially distort the pretrained representation.

To stabilize pseudo-label generation, we employ a student-teacher framework with identical architectures. We maintain a student model $f_\theta$ and a teacher model $f_{\theta'}$, where the teacher parameters are updated as an exponential moving average of the student parameters:
\begin{equation}
\theta' \leftarrow \mu \theta' + (1-\mu)\theta,
\end{equation}
where $\mu\in[0,1)$ controls temporal smoothing. By aggregating historical student states, the teacher produces stabilized predictions that reduce pseudo-label variance. These predictions are used for progressive subset selection and consistency learning. We use the teacher only as a temporally smoothed target generator; gradients are applied to the student model.

\subsection{Progressive Margin-Based Subset Selection}

A central challenge in unsupervised adaptation is pseudo-label reliability. Since target labels are unavailable, training signals come from model predictions, which can be noisy, especially early in adaptation. Using all pseudo-labeled samples can introduce substantial label noise, leading to confirmation bias and unstable optimization. This issue is particularly pronounced for PVLMs under domain shift, where visual representations may deviate from the pretrained embedding space.

To mitigate this problem, we construct a subset of high-confidence pseudo-labeled samples and use it as supervised anchors. Rather than selecting samples solely by predicted probability, we rank them by prediction margin, which measures the separation between the most likely and competing classes. The motivation is that the maximum probability can be poorly calibrated under domain shift, whereas the margin directly measures the relative separation between the top predicted class and its nearest competitor in the prototype classifier. Given an unlabeled image $\mathbf{x}$, the teacher model produces logits
\begin{equation}
\mathbf{z} = f_{\theta'}(\mathbf{x}) \in \mathbb{R}^{C},
\end{equation}
and the pseudo-label is
\begin{equation}
\hat{y} = \arg\max_{c} z_c .
\end{equation}

We define the prediction margin as
\begin{equation}
m(\mathbf{x}) = z_{(1)} - z_{(2)},
\end{equation}
where $z_{(1)}$ and $z_{(2)}$ are the largest and second-largest logits. Since the classifier uses cosine similarity between normalized image features and class prototypes, the margin reflects separation between competing prototypes; larger margins typically indicate more reliable pseudo-labels. We do not assume that large margins guarantee correctness, especially under severe domain shift. Instead, we use margins as a conservative ranking criterion and combine them with class-balanced selection and progressive expansion to reduce the effect of overconfident errors.

Margin-based ranking alone can yield imbalanced supervision across classes. If some classes produce consistently higher-confidence predictions, they may dominate the selected subset and bias adaptation. We therefore enforce class-balanced selection. For each class $c$, we collect samples with $\hat{y}=c$, rank them by $m(\mathbf{x})$, and select the top $K_e$ at epoch $e$. To reduce early error propagation, we progressively expand the subset size:
\begin{equation}
K_e = \min(K_{\max}, K_0 + e \Delta K),
\end{equation}
where $K_0$ is the initial subset size, $\Delta K$ is the per-epoch increment, and $K_{\max}$ is the maximum. This schedule restricts early training to the most confident samples and increases coverage as pseudo-label quality improves. If fewer than $K_e$ samples are predicted as class $c$, all available samples for that class are selected. This avoids forcing class balance when the teacher does not predict enough samples for a class.

Formally, the selected pseudo-labeled subset for class $c$ is
\begin{equation}
\mathcal{S}_e^{(c)}
=
\operatorname{TopK}_{K_e}
\left(
\left\{ (\mathbf{x}, \hat{y}) \mid \hat{y}=c \right\};
m(\mathbf{x})
\right),
\end{equation}
and the overall subset is
\begin{equation}
\mathcal{S}_e
=
\bigcup_{c=1}^{C} \mathcal{S}_e^{(c)}.
\end{equation}

The student is trained on $\mathcal{S}_e$ using a supervised classification loss. We denote by $\ell(\mathbf{z},y)$ the per-sample loss, which keeps the objective compatible with different PVLM architectures (cross-entropy for CLIP, binary cross-entropy for SigLIP):
\begin{equation}
\mathcal{L}_{\text{sup}}
=
\frac{1}{|\mathcal{S}_e|}
\sum_{(\mathbf{x},\hat{y}) \in \mathcal{S}_e}
\ell\!\left(f_\theta(\mathbf{x}), \hat{y}\right).
\end{equation}

Importantly, pseudo-labels for subset construction are generated by the teacher rather than the student. Since $\theta'$ is an EMA of $\theta$, the teacher predictions are typically more stable, improving both margin estimation and subset selection. Overall, progressive margin-based subset selection provides reliable supervision anchors while controlling error propagation. This component is intended to control which pseudo-labels are trusted for direct supervised training, rather than to use every target sample as if it were labeled.

\subsection{Cascaded Consistency Learning}

Progressive subset selection provides reliable anchors but covers only a small portion of the target data, especially early in adaptation. To leverage the full unlabeled dataset $\mathcal{U}$, we enforce cascaded consistency on all samples using a dynamic confidence threshold to filter unreliable teacher pseudo-labels. While consistency learning encourages prediction invariance under perturbations, single-stage consistency based on student predictions can amplify pseudo-label noise and destabilize training. This is particularly problematic in unsupervised adaptation, where labeled validation data is unavailable for early stopping.

We introduce \emph{cascaded consistency learning}, which integrates temporal stabilization from the teacher with cross-view consistency across stochastic augmentations. The term ``cascaded'' refers to two sequential supervision stages: first, the EMA teacher produces a temporally smoothed target from a weakly augmented view; second, this target supervises the student on a strongly augmented view. This separates target generation from student optimization and avoids using instantaneous student predictions as their own direct targets. The key idea is to generate pseudo-labels using teacher predictions on weak views and supervise student predictions on strong views, which reduces pseudo-label variance and stabilizes adaptation.

Given $\mathbf{x} \in \mathcal{U}$, we generate a weak view $\mathbf{x}^{w}$ and a strong view $\mathbf{x}^{s}$. The weak view uses standard evaluation-style preprocessing with mild stochastic augmentation, while the strong view applies stronger image perturbations following the FixMatch-style augmentation protocol described in the implementation details. The teacher produces per-class probabilities from the weak view,
\begin{equation}
\mathbf{p}^{t} = g(f_{\theta'}(\mathbf{x}^{w})) \in [0,1]^C,
\end{equation}
where $g(\mathbf{z})=\mathrm{softmax}(\mathbf{z})$ for CLIP and $g(\mathbf{z})=\mathrm{sigmoid}(\mathbf{z})$ for SigLIP. The teacher pseudo-label is
\begin{equation}
\hat{y}^{t} = \arg\max_{c} p^{t}_c ,
\end{equation}
and we measure confidence by
\begin{equation}
\gamma(\mathbf{x}) = \max_{c} p^{t}_c .
\end{equation}

To avoid enforcing unreliable supervision, we introduce a dynamic confidence threshold $\tau_e$ and retain pseudo-labels only when $\gamma(\mathbf{x})\ge\tau_e$. We use a progressively relaxed schedule: early in training we apply a stricter threshold to suppress noisy pseudo-labels, and later relax it as the model adapts. Specifically,
\begin{equation}
\tau_e =
\tau_{\text{start}}
+
\min\!\left(1,\frac{e}{E_{\text{warm}}}\right)
\left(\tau_{\text{end}}-\tau_{\text{start}}\right),
\end{equation}
where $\tau_{\text{start}}$ and $\tau_{\text{end}}$ are the initial and final thresholds and $E_{\text{warm}}$ is the linear annealing length. This complements progressive subset expansion by gradually incorporating more supervision as pseudo-label reliability improves. In our implementation, $\tau_{\text{start}} > \tau_{\text{end}}$, so the threshold is relaxed over training. This conservative-to-permissive schedule is designed to suppress noisy consistency targets in early epochs while increasing unlabeled-data usage later.

The student produces logits from the strong view $f_{\theta}(\mathbf{x}^{s})\in\mathbb{R}^{C}$. We enforce weak-to-strong agreement for all $\mathbf{x}\in\mathcal{U}$ (including samples in $\mathcal{S}_e$) via
\begin{equation}
\mathcal{L}_{\mathrm{cons}}
=
\mathbb{E}_{\mathbf{x}\sim\mathcal{U}}
\mathbbm{1}\!\left[\gamma(\mathbf{x}) \ge \tau_e\right]
\,
\ell(f_{\theta}(\mathbf{x}^{s}), \hat{y}^{t}).
\end{equation}
Because the teacher is an EMA of the student, its predictions are more stable than instantaneous student outputs. Compared to single-stage consistency, the proposed cascade reduces pseudo-label variance and mitigates oscillatory training behavior. The supervised subset loss and the consistency loss play different roles: $\mathcal{L}{\mathrm{sup}}$ provides direct optimization on selected high-margin anchors, while $\mathcal{L}{\mathrm{cons}}$ regularizes predictions over the broader unlabeled set.

The overall training objective is
\begin{equation}\label{eq:overall}
\min_{\theta}
\;
\mathcal{L}
=
\mathcal{L}_{\mathrm{sup}}
+
\lambda\,\mathcal{L}_{\mathrm{cons}},
\end{equation}
where $\lambda$ controls the strength of consistency regularization. Together with progressive subset selection, cascaded consistency forms a self-reinforcing loop: reliable anchors stabilize early adaptation, and consistency learning exploits all unlabeled data and improves pseudo-label quality over time. We keep $\lambda$ fixed across datasets to avoid dataset-specific tuning in the unlabeled setting.




At each epoch, the EMA teacher evaluates all samples in $\mathcal{U}$ and constructs a class-balanced pseudo-labeled subset $\mathcal{S}_e$ using progressive margin-based selection, where the per-class subset size $K_e$ increases over training. The student is then optimized using Equation~\ref{eq:overall}, with mini-batches from $\mathcal{S}_e$ providing supervised signals and mini-batches from $\mathcal{U}$ enforcing confidence filtered weak-to-strong consistency with threshold $\tau_e$. After each update, the teacher parameters are updated by exponential moving average. The overall procedure is summarized in Algorithm~\ref{alg:method} in the appendix~\ref{sec:algo}. This epoch-level subset construction requires teacher inference over the unlabeled target set. We report this cost in the implementation and efficiency analysis rather than treating the method as overhead-free.


\section{Experiments}
\label{sec:exp}

\subsection{Experimental Setup}

\paragraph{Datasets.}
We evaluate the proposed method on several widely used benchmark datasets for vision–language model evaluation, including DTD \citep{cimpoi2014describing}, FGVCAircraft \citep{maji2013fine}, OxfordPets \citep{parkhi2012cats}, CIFAR10 \citep{krizhevsky2009learning}, CIFAR100 \citep{krizhevsky2009learning}, and Food101 \citep{bossard2014food}. These datasets cover diverse visual recognition tasks such as fine-grained classification, texture recognition, and object recognition. Importantly, they exhibit varying levels of zero-shot performance for pretrained vision–language models, ranging from low-accuracy datasets (FGVCAircraft and DTD) to medium-accuracy datasets (CIFAR100 and Food101) and high-accuracy datasets (OxfordPets and CIFAR10). 
Detailed dataset statistics and official split information are provided in the supplementary material.

Following the common evaluation protocol for pretrained vision-language models, all target datasets are treated as \emph{fully unlabeled} during adaptation. No labeled target data is used for training or model selection, and the test sets are used only for final evaluation.

\paragraph{PVLMs.}
We evaluate our method using several pretrained vision-language models, including CLIP \citep{radford2021learning} and SigLIP \citep{zhai2023sigmoid} architectures. In particular, experiments are conducted using CLIP ViT-B/32, CLIP ViT-B/16, and SigLIP ViT-B/16 models to verify the generality of the proposed framework across different PVLM architectures. Unlike CLIP, SigLIP is trained with a sigmoid-based contrastive objective, so results on SigLIP help test whether the proposed method depends on CLIP-specific logit behavior.

For each dataset, class descriptions are constructed using multiple hand-written textual prompts from CuPL \citep{pratt2023does}, and class prototypes are initialized by averaging normalized embeddings of prompts corresponding to each class. The same prompt set is used for the zero-shot baseline and for our adapted model, so the reported gains over zero-shot reflect the effect of adaptation rather than a change in prompting. Since some comparison results are taken from prior work, differences in prompt sets or implementation details may affect direct cross-method comparison; we therefore interpret comparisons with published baselines as indicative rather than fully controlled.

\paragraph{Baselines.}
We compare our method with zero-shot PVLM baselines and representative unsupervised adaptation methods. \textbf{CLIP} \citep{radford2021learning} and \textbf{SigLIP} \citep{zhai2023sigmoid} directly apply pretrained vision--language models with textual prompts and no target-domain adaptation. \textbf{UPL} \citep{huang2022unsupervised}, \textbf{POUF} \citep{tanwisuth2023pouf}, \textbf{LaFTer} \citep{mirza2023lafter}, and \textbf{DPA} \citep{ali2025dpa} adapt PVLMs using unlabeled target data and serve as representative unsupervised adaptation baselines. Implementation details are attached in the appendix \ref{sec:imple-detail}


\subsection{Main Results}

\begin{table}[t]
\centering
\caption{Comparison with existing unsupervised PVLM adaptation methods using \textbf{CLIP ViT-B/32}. Accuracy (\%) on six benchmark datasets. Baseline adaptation results are taken from published DPA comparisons.}
\label{tab:main_vitb32}
\resizebox{\textwidth}{!}{
\begin{tabular}{l|cccccc}
\toprule
Methods & FGVCAircraft & DTD & CIFAR100 & Food101 & OxfordPets & CIFAR10 \\
\midrule
CLIP \citep{radford2021learning} & 18.9 & 44.4 & 62.3 & 83.3 & 87.5 & 88.3 \\
UPL \citep{huang2022unsupervised} & 17.1 & 45.4 & 67.4 & 84.3 & 83.8 & 91.3 \\
POUF \citep{tanwisuth2023pouf} & 19.8 & 47.0 & 68.9 & 84.0 & 89.5 & 92.5 \\
LaFTer \citep{mirza2023lafter} & 19.9 & 50.3 & 69.8 & 82.5 & 84.9 & 94.6 \\
DPA \citep{ali2025dpa} & \underline{20.7} & \underline{55.7} & \textbf{76.5} & \textbf{84.8} & \textbf{90.7} & \textbf{96.0} \\
\textbf{Ours} & \textbf{22.9} & \textbf{57.7} & \underline{74.1} & \textbf{84.8} & \underline{90.3} & \textbf{96.0} \\
\bottomrule
\end{tabular}
}
\end{table}

\begin{table}[t]
\centering
\caption{Comparison with existing unsupervised PVLM adaptation methods using \textbf{CLIP ViT-B/16}. Accuracy (\%) on the same six benchmark datasets. Food101 for ViT-B/16 is not reported by some baselines}
\label{tab:main_vitb16}
\resizebox{\textwidth}{!}{
\begin{tabular}{l|cccccc}
\toprule
Methods & FGVCAircraft & DTD & CIFAR100 & Food101 & OxfordPets & CIFAR10 \\
\midrule
CLIP \citep{radford2021learning} & 24.9 & 43.9 & 66.6 & 87.7 & 89.0 & 90.1 \\
UPL \citep{huang2022unsupervised} & 22.5 & 45.9 & 70.4 & -- & 88.0 & 92.9 \\
POUF \citep{tanwisuth2023pouf} & 24.4 & 48.6 & 66.3 & -- & 91.8 & 92.3 \\
LaFTer \citep{mirza2023lafter} & 22.4 & \textbf{54.8} & \underline{76.3} & -- & 85.3 & 96.5 \\
DPA \citep{ali2025dpa} & \underline{25.1} & 50.7 & \textbf{78.1} & -- & \underline{93.2} & 96.4 \\
\textbf{Ours} & \textbf{26.2} & \underline{53.4} & 76.1 & \textbf{89.3} & \textbf{93.5} & \textbf{97.2} \\
\bottomrule
\end{tabular}
}
\end{table}

We compare the proposed method with several representative approaches for unsupervised adaptation of pretrained vision-language models, including UPL, POUF, LaFTer, and DPA. Classification accuracy is reported on six benchmark datasets using CLIP ViT-B/32 and CLIP ViT-B/16 models. The results are summarized in Table~\ref{tab:main_vitb32} and Table~\ref{tab:main_vitb16}. Following common practice, results for the comparison methods are taken directly from the DPA paper to ensure fair and consistent evaluation.

As shown in Table~\ref{tab:main_vitb32} and Table~\ref{tab:main_vitb16}, the proposed method consistently improves over the CLIP zero-shot baseline. The gains are especially clear on datasets with lower zero-shot accuracy, such as FGVCAircraft and DTD. For example, with CLIP ViT-B/32, our method improves FGVCAircraft from 18.9\% to 22.9\% and DTD from 44.4\% to 57.7\%, suggesting that progressive subset selection and cascaded consistency are effective under larger domain shifts.

Compared with existing unsupervised adaptation methods, our approach achieves competitive or superior performance on most datasets. For CLIP ViT-B/32, it obtains the best results on FGVCAircraft and DTD and matches the best results on Food101 and CIFAR10. For CLIP ViT-B/16, it achieves the highest accuracy on FGVCAircraft, Food101, OxfordPets, and CIFAR10. However, it does not uniformly outperform DPA; DPA remains stronger on CIFAR100 for both CLIP backbones and on DTD for ViT-B/16. This suggests that distribution-aware prototype alignment may be beneficial when many semantically related classes produce ambiguous pseudo-labels.

Overall, the results show that the proposed method provides a simple and effective strategy for unsupervised PVLM adaptation. It consistently improves the corresponding zero-shot models and remains competitive with recent unsupervised adaptation methods, while also revealing cases where stronger pseudo-label calibration or distribution-aware alignment may further improve performance.

\subsection{Generalization across PVLM Architectures}

In addition to the CLIP-based experiments, we evaluate the proposed method on SigLIP ViT-B/16 \citep{zhai2023sigmoid} to verify whether the framework generalizes across PVLM architectures. SigLIP is trained using a sigmoid-based contrastive objective and exhibits different logit characteristics compared with CLIP. Evaluating on SigLIP therefore tests whether the proposed adaptation framework is limited to CLIP-style contrastive models.

\begin{table}[t]
\centering
\caption{Additional results using SigLIP ViT-B/16. Accuracy (\%) on six benchmark datasets.}
\label{tab:siglip_results}
\resizebox{\textwidth}{!}{
\begin{tabular}{l|cccccc|c}
\toprule
Method & FGVCAircraft & DTD & CIFAR100 & Food101 & OxfordPets & CIFAR10 & Avg \\
\midrule
SigLIP \citep{zhai2023sigmoid} & 43.9 & 63.4 & 70.8 & 89.4 & 94.2 & 92.5 & 75.7 \\
Ours & 47.4 & 67.9 & 82.9 & 91.6 & 95.2 & 97.6 & \textbf{80.4} (+4.7) \\
\bottomrule
\end{tabular}
}
\end{table}

As shown in Table~\ref{tab:siglip_results}, the proposed method improves SigLIP on all six datasets, increasing the average accuracy from 75.7\% to 80.4\% (+4.7 points). The gains are especially large on CIFAR100 (+12.1), CIFAR10 (+5.1), DTD (+4.5), and FGVCAircraft (+3.5). These results suggest that the proposed framework is not limited to CLIP, since it can also adapt a PVLM trained with a different image-text objective. We emphasize that this experiment is intended to evaluate cross-architecture applicability relative to the corresponding SigLIP baseline; it is not a controlled comparison with prior unsupervised adaptation methods on SigLIP. More results about ablation study and training stability analysis are included in the appendix \ref{sec:ablation} and \ref{sec:train-stab}.


\section{Conclusion}

We presented a simple framework for unsupervised adaptation of pretrained vision--language models by combining progressive margin-based subset selection with cascaded consistency learning. The method selects class-balanced high-margin pseudo-labeled samples as supervision anchors, gradually expands the selected subset, and uses confidence-filtered EMA teacher predictions to exploit unlabeled target data through weak-to-strong consistency. By updating only normalization parameters and class prototypes, the framework preserves most pretrained parameters while adapting to target domains. Experiments across multiple benchmarks with CLIP and SigLIP show consistent improvements over zero-shot baselines and competitive performance against recent unsupervised PVLM adaptation methods. Future work will explore broader benchmarks, controlled comparisons with more recent baselines, and adaptive schedules for pseudo-label selection and confidence filtering.

\section*{Acknowledgment}
This work was partly supported by the Natural Sciences and
Engineering Research Council of Canada (NSERC) Discovery Grants and DND/NSERC Discovery Grant Supplements under grant nos RGPIN-2026-05645 and DGDND-2026-05645.

%
%
\bibliographystyle{plainnat}
\bibliography{ref}


\newpage
\appendix
\section{Appendix}

\subsection{Algorithm}\label{sec:algo}
The overall procedure is summarized in Algorithm~\ref{alg:method}.

\begin{algorithm}[ht!]
\caption{Progressive Subset Selection with Cascaded Consistency}
\label{alg:method}
\KwIn{Unlabeled dataset $\mathcal{U}$, training epochs $E$, initial subset size $K_0$, subset increment $\Delta K$, maximum subset size $K_{\max}$, threshold schedule $(\tau_{\text{start}}, \tau_{\text{end}}, E_{\text{warm}})$, consistency weight $\lambda$, EMA momentum $\mu$}
\KwOut{Adapted student model $f_\theta$}

Initialize student model $f_\theta$ and teacher model $f_{\theta'} \leftarrow f_\theta$\;

\For{epoch $e = 0,\dots,E-1$}{
    Update subset size $K_e \leftarrow \min(K_{\max}, K_0 + e\Delta K)$\;
    Compute $\tau_e$ from the threshold schedule\;

    \tcp{Progressive subset selection using the teacher}
    Compute teacher logits $\mathbf{z}=f_{\theta'}(\mathbf{x})$ for $\mathbf{x}\in\mathcal{U}$\;
    Compute pseudo-label $\hat{y}=\arg\max_c z_c$ and margin $m(\mathbf{x})$ for each $\mathbf{x}\in\mathcal{U}$\;
    Select class-balanced pseudo-labeled subset $\mathcal{S}_e$ via Top-$K_e$ margin ranking\;

    \For{each training iteration}{
        Sample a labeled batch $\{(\mathbf{x}_i,\hat{y}_i)\}$ from $\mathcal{S}_e$\;
        Sample an unlabeled batch $\{\mathbf{x}_j\}$ from $\mathcal{U}$\;

        \tcp{Supervised update on pseudo-labeled anchors}
        Compute supervised loss $\mathcal{L}_{\mathrm{sup}}$ on $\{(\mathbf{x}_i,\hat{y}_i)\}$\;

        \tcp{Cascaded consistency: teacher weak $\rightarrow$ student strong}
        Generate weak and strong augmentations $(\mathbf{x}_j^{w}, \mathbf{x}_j^{s})$\;
        Compute teacher probabilities $\mathbf{p}^{t}_j=g(f_{\theta'}(\mathbf{x}_j^{w}))$\;
        Obtain teacher pseudo-label $\hat{y}^{t}_j=\arg\max_c p^{t}_{j,c}$ and confidence $\gamma_j=\max_c p^{t}_{j,c}$\;
        Compute consistency loss $\mathcal{L}_{\mathrm{cons}}$ using $\mathbbm{1}[\gamma_j\ge\tau_e]\ell(f_{\theta}(\mathbf{x}_j^{s}),\hat{y}^{t}_j)$\;

        \tcp{Optimize student and update teacher}
        Update student parameters $\theta$ using $\mathcal{L}=\mathcal{L}_{\mathrm{sup}}+\lambda\mathcal{L}_{\mathrm{cons}}$\;
        Update teacher parameters $\theta' \leftarrow \mu\theta' + (1-\mu)\theta$\;
    }
}
\Return adapted model $f_\theta$
\end{algorithm}

\subsection{Implementation Details}\label{sec:imple-detail}

We adopt pretrained CLIP and SigLIP models from the HuggingFace Transformers library \citep{wolf2020transformers}. After initializing class prototypes using textual prompts, the textual encoder is discarded and a cosine classifier is constructed on top of the visual encoder.

To ensure stable and efficient adaptation, we follow a lightweight training strategy. Specifically, only the affine parameters of normalization layers in the visual encoder and the class prototypes are updated, while all other parameters of the pretrained model remain frozen. The logit scale parameter inherited from the pretrained model is also kept fixed. 
%
For progressive subset selection, the initial number of selected samples per class is set to $K_0 = 5$, and the subset size increases by $\Delta K = 5$ per epoch until reaching a maximum of $K_{\max} = 100$. Pseudo-labels are generated using the teacher model, and samples are ranked based on prediction margins within each class.
For cascaded consistency learning, the confidence threshold is scheduled linearly from $\tau_{\text{start}} = 0.98$ to $\tau_{\text{end}} = 0.95$ during the first $E_{\text{warm}} = 10$ epochs, after which the threshold remains fixed. The consistency loss weight is set to $\lambda = 1.0$. The consistency weight is fixed across datasets for the same reason as the subset schedule: no labeled validation data is assumed for dataset-specific tuning.

Training is performed for $E = 50$ epochs using the AdamW optimizer with a learning rate of $1\times10^{-4}$. The exponential moving average momentum used for updating the teacher model is set to $\mu = 0.99$. The batch size for unlabeled samples is 128 and the batch size for pseudo-labeled samples is 16. During training, weak and strong augmentations follow the standard FixMatch \citep{sohn2020fixmatch} augmentation strategy.
All hyperparameters are selected empirically using simple settings and are kept fixed across all datasets. Since unsupervised adaptation does not provide labeled validation data for model selection, we avoid extensive hyperparameter tuning and instead adopt straightforward configurations that work consistently across different tasks.

\subsection{Ablation Study}\label{sec:ablation}

To better understand the contributions of key components in our framework, we conduct ablation studies on the subset expansion schedule and the confidence threshold schedule. Results are reported on DTD, FGVCAircraft, and OxfordPets using CLIP ViT-B/32. The results are summarized in Table~\ref{tab:ablation}.

\begin{table}[t]
\centering
\caption{Ablation study on the progressive subset expansion schedule ($K$) and the confidence threshold schedule ($\tau$) in cascaded consistency learning. Accuracy is reported in \%.}
\label{tab:ablation}
\begin{tabular}{l|l|ccc}
\toprule
Subset Schedule ($K$) & Threshold ($\tau$) & FGVCAircraft & DTD & OxfordPets \\
\midrule
Fixed & $0.95$ & 21.4 & 55.8 & 89.3 \\
Progressive & $0.95$ & 21.3 & 57.1 & 89.4 \\
Progressive & $0.95 \rightarrow 0.90$ & 21.8 & \textbf{58.4} & 89.9 \\
Progressive & $0.98 \rightarrow 0.95$ & \textbf{22.9} & 57.7 & \textbf{90.3} \\
\bottomrule
\end{tabular}
\end{table}

\paragraph{Effect of progressive subset expansion.}
We first evaluate the effect of the subset expansion strategy used in progressive margin-based subset selection. Specifically, we compare a fixed subset size with a progressive schedule that gradually increases the number of selected samples per class during training.
Using a progressive schedule consistently improves performance compared with using a fixed subset size. For example, on DTD, the progressive schedule improves accuracy from 55.8\% to 57.1\%, demonstrating that progressively expanding the pseudo-labeled subset allows the model to incorporate more reliable supervision as training proceeds. This observation supports the design motivation of progressive subset selection: early training should rely on a small set of high-confidence samples, while later stages can safely incorporate additional pseudo-labeled data.

\paragraph{Effect of the confidence threshold schedule.}
We further study the impact of the confidence threshold used in cascaded consistency learning. Instead of using a fixed threshold, we evaluate schedules that gradually relax the threshold during training.
Introducing a threshold schedule leads to consistent improvements. For instance, using a schedule from $0.95$ to $0.90$ improves DTD performance from 57.1\% to 58.4\%, indicating that progressively incorporating more unlabeled samples helps improve representation adaptation once the model becomes more reliable. We also observe that a stricter initial threshold ($0.98 \rightarrow 0.95$) leads to the best performance on FGVCAircraft and OxfordPets, suggesting that starting with highly confident pseudo-labels is beneficial for fine-grained datasets where early pseudo-label noise can significantly affect training stability.


\subsection{Training Stability Analysis}\label{sec:train-stab}

Unsupervised adaptation lacks labeled validation data for model selection or early stopping, making training stability an important practical consideration. We therefore analyze the training dynamics of the proposed method by tracking test accuracy across adaptation epochs. This analysis is used only to understand optimization behavior; test labels are not used for training, checkpoint selection, or hyperparameter tuning.

\begin{figure}[t]
\centering
\includegraphics[width=\linewidth]{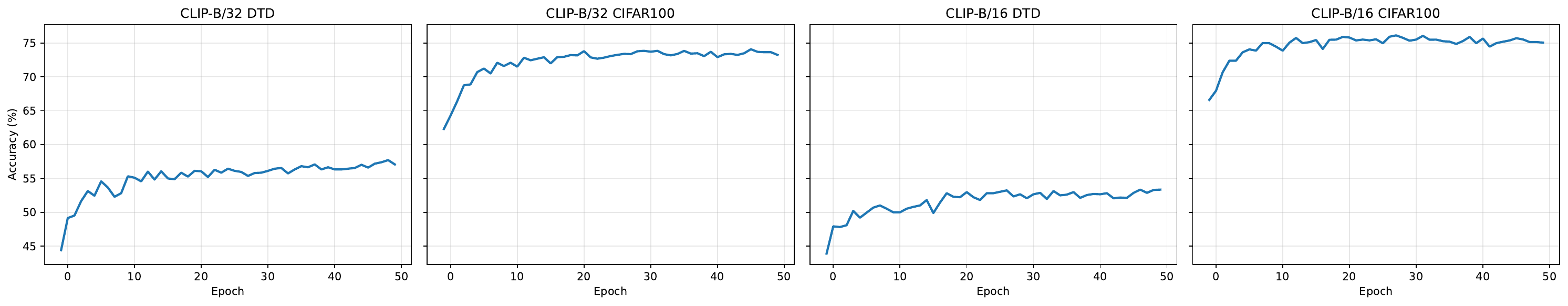}
\caption{Training stability of the proposed method. Test accuracy is shown across training epochs starting from the zero-shot initialization for CLIP ViT-B/32 and CLIP ViT-B/16 on DTD and CIFAR100.}
\label{fig:training_stability}
\end{figure}

Figure~\ref{fig:training_stability} shows the training curves for CLIP ViT-B/32 and CLIP ViT-B/16 on DTD and CIFAR100. Across these settings, the proposed method improves over the zero-shot initialization and exhibits smooth convergence without severe late-stage degradation. The gains are more pronounced on DTD, where the zero-shot performance is relatively low, suggesting that progressive subset selection can provide useful supervision anchors when the initial classifier is imperfect but still contains reliable high-margin predictions. On CIFAR100, the improvements are more moderate, which is consistent with the main results and suggests that datasets with many semantically related classes may remain challenging for margin-based pseudo-label selection.

\begin{figure}[t]
\centering
\includegraphics[width=0.75\linewidth]{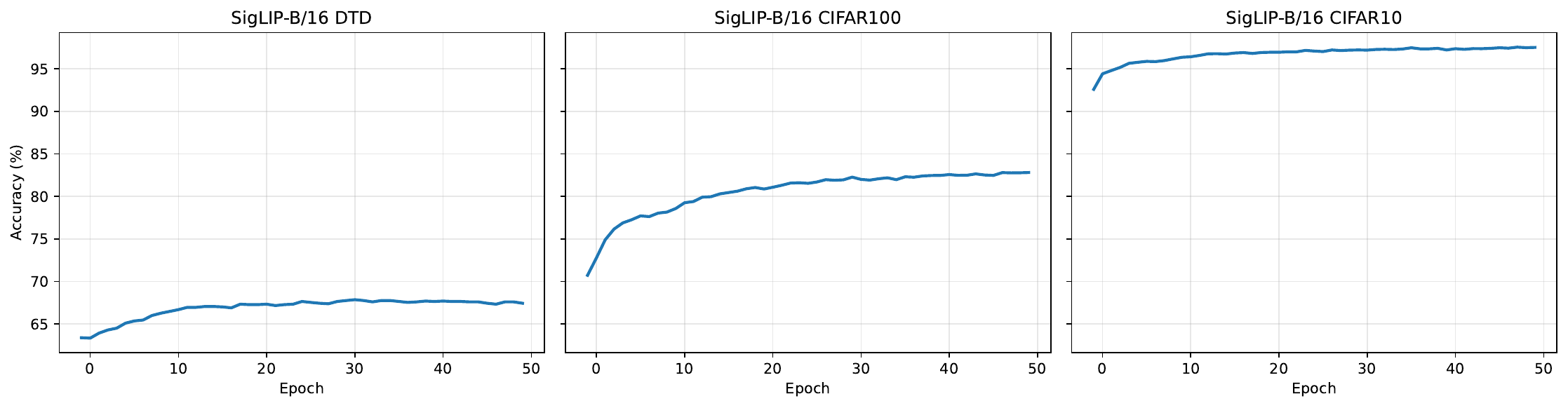}
\caption{Training stability of the proposed method using SigLIP ViT-B/16. Test accuracy across training epochs starting from the zero-shot initialization on DTD, CIFAR100, and CIFAR10.}
\label{fig:siglip_training_stability}
\end{figure}

Figure~\ref{fig:siglip_training_stability} further reports training curves for SigLIP ViT-B/16 on DTD, CIFAR100, and CIFAR10. Similar to the CLIP results, the adapted model improves over the zero-shot initialization and remains stable throughout training. This indicates that the observed training behavior is not limited to CLIP-style contrastive models, but also appears when adapting SigLIP, which uses a different image--text training objective.

Overall, these results suggest that the proposed combination of progressive subset selection and cascaded consistency can produce stable adaptation trajectories across different PVLM backbones. However, because the curves only report our method, they should be interpreted as evidence that the proposed training procedure itself does not suffer from severe oscillation or late-stage degradation in the evaluated settings. They do not establish that our method is more stable than all prior adaptation methods. A stronger comparative stability analysis would require baseline training curves, multi-seed error bands, and best-versus-final checkpoint gaps.


\newpage
\section*{NeurIPS Paper Checklist}

\begin{enumerate}

\item {\bf Claims}
    \item[] Question: Do the main claims made in the abstract and introduction accurately reflect the paper's contributions and scope?
    \item[] Answer: \answerYes{} 
    \item[] Justification: The main claims made in the abstract and introduction accurately reflect the paper's contributions and scope.

\item {\bf Limitations}
    \item[] Question: Does the paper discuss the limitations of the work performed by the authors?
    \item[] Answer: \answerYes{} 
    \item[] Justification: The paper discussed the limitations of the work performed by the authors.

\item {\bf Theory assumptions and proofs}
    \item[] Question: For each theoretical result, does the paper provide the full set of assumptions and a complete (and correct) proof?
    \item[] Answer: \answerNA{} 
    \item[] Justification: The paper does not include theoretical results.

    \item {\bf Experimental result reproducibility}
    \item[] Question: Does the paper fully disclose all the information needed to reproduce the main experimental results of the paper to the extent that it affects the main claims and/or conclusions of the paper (regardless of whether the code and data are provided or not)?
    \item[] Answer: \answerYes{} 
    \item[] Justification: The paper fully disclose all the information needed to reproduce the main experimental results of the paper.

\item {\bf Open access to data and code}
    \item[] Question: Does the paper provide open access to the data and code, with sufficient instructions to faithfully reproduce the main experimental results, as described in supplemental material?
    \item[] Answer: \answerYes{} 
    \item[] Justification: The data is public and the code will be made public upon acceptance.

\item {\bf Experimental setting/details}
    \item[] Question: Does the paper specify all the training and test details (e.g., data splits, hyperparameters, how they were chosen, type of optimizer) necessary to understand the results?
    \item[] Answer: \answerYes{} 
    \item[] Justification: The paper specified all the training and test details.

\item {\bf Experiment statistical significance}
    \item[] Question: Does the paper report error bars suitably and correctly defined or other appropriate information about the statistical significance of the experiments?
    \item[] Answer: \answerYes{} 
    \item[] Justification: The experimental results are averaged over 3 runs with different random seeds.

\item {\bf Experiments compute resources}
    \item[] Question: For each experiment, does the paper provide sufficient information on the computer resources (type of compute workers, memory, time of execution) needed to reproduce the experiments?
    \item[] Answer: \answerYes{} 
    \item[] Justification: The experiments can be conducted on standard GPU, for example, NVIDIA RTX 3090 or H100.
    
\item {\bf Code of ethics}
    \item[] Question: Does the research conducted in the paper conform, in every respect, with the NeurIPS Code of Ethics \url{https://neurips.cc/public/EthicsGuidelines}?
    \item[] Answer: \answerYes{} 
    \item[] Justification: The research conducted in the paper conforms, in every respect, with the NeurIPS Code of Ethics.

\item {\bf Broader impacts}
    \item[] Question: Does the paper discuss both potential positive societal impacts and negative societal impacts of the work performed?
    \item[] Answer: \answerYes{} 
    \item[] Justification: The paper discussed both potential positive societal impacts and negative societal impacts of the work performed.
    
\item {\bf Safeguards}
    \item[] Question: Does the paper describe safeguards that have been put in place for responsible release of data or models that have a high risk for misuse (e.g., pre-trained language models, image generators, or scraped datasets)?
    \item[] Answer: \answerNA{} 
    \item[] Justification: \answerNA{}

\item {\bf Licenses for existing assets}
    \item[] Question: Are the creators or original owners of assets (e.g., code, data, models), used in the paper, properly credited and are the license and terms of use explicitly mentioned and properly respected?
    \item[] Answer: \answerYes{} 
    \item[] Justification: The creators or original owners of assets (e.g., code, data, models), used in the paper, are properly credited and the license and terms of use are explicitly mentioned and properly respected.

\item {\bf New assets}
    \item[] Question: Are new assets introduced in the paper well documented and is the documentation provided alongside the assets?
    \item[] Answer: \answerNA{} 
    \item[] Justification: The paper does not release new assets.

\item {\bf Crowdsourcing and research with human subjects}
    \item[] Question: For crowdsourcing experiments and research with human subjects, does the paper include the full text of instructions given to participants and screenshots, if applicable, as well as details about compensation (if any)? 
    \item[] Answer: \answerNA{} 
    \item[] Justification: The paper does not involve crowdsourcing nor research with human subjects.

\item {\bf Institutional review board (IRB) approvals or equivalent for research with human subjects}
    \item[] Question: Does the paper describe potential risks incurred by study participants, whether such risks were disclosed to the subjects, and whether Institutional Review Board (IRB) approvals (or an equivalent approval/review based on the requirements of your country or institution) were obtained?
    \item[] Answer: \answerNA{} 
    \item[] Justification: The paper does not involve crowdsourcing nor research with human subjects.

\item {\bf Declaration of LLM usage}
    \item[] Question: Does the paper describe the usage of LLMs if it is an important, original, or non-standard component of the core methods in this research? Note that if the LLM is used only for writing, editing, or formatting purposes and does \emph{not} impact the core methodology, scientific rigor, or originality of the research, declaration is not required.
    \item[] Answer: \answerNA{} 
    \item[] Justification: LLMs are used only for writing, editing, or formatting purposes.

\end{enumerate}

\end{document}